%% file: main.tex
\documentclass[conference,10pt,nofonttune,dvipsnames,hidelinks]{IEEEtran}
\input{head/header.tex}

\input{acronyms}
\definecolor{lightcyan}{rgb}{0.84,1,1}
\definecolor{lightgreen}{rgb}{0.64,1,0.71}
\definecolor{lightred}{rgb}{1,0.7,0.71}

\usepackage{array}

\definecolor{lightblue}{rgb}{0.8, 0.8, 1.0} 
\definecolor{lightgreen}{rgb}{0.8, 1.0, 0.8} 
\definecolor{pinkish}{rgb}{1.0, 0.8, 0.8} 

\tikzstyle{startstop} = [rectangle, rounded corners, text centered, draw=black, fill=red!30]
\tikzstyle{io} = [trapezium, trapezium left angle=70, trapezium right angle=110, text centered, draw=black, fill=blue!30]
\tikzstyle{process} = [rectangle, text centered, draw=black, fill=orange!30]
\tikzstyle{decision} = [diamond, aspect=2, text centered, draw=black, fill=green!30]
\tikzstyle{arrow} = [thick,->,>=stealth]

\usetikzlibrary{positioning, fit, shapes, backgrounds}

\newif\ifuseicons
\useiconstrue

\newcommand{\CopyrightNotice}[2]{%
  \begin{picture}(0,0)(0,0)
    \put(#1){\parbox{\paperwidth}{\sf \center {\small%
      This is the author's version of the work. The definitive work was published in the \emph{Proceedings of the\\
      International Conference on Field Programmable Technology (FPT), Sydney, Australia, December 10--12, 2024.} \copyright\ IEEE, 2024}}}%
  \end{picture}
  \vspace{#2}
}

\begin{document}

\title{Hardware/Software Co-Design of RISC-V Extensions for Accelerating Sparse DNNs on FPGAs}
\title{\fontsize{23.8}{28}\selectfont Hardware/Software Co-Design of RISC-V Extensions for Accelerating Sparse DNNs on FPGAs\\[-2mm]}

\author{%
\IEEEauthorblockN{Muhammad Sabih, Abrarul Karim, Jakob Wittmann, Frank Hannig, and Jürgen Teich}
\IEEEauthorblockA{Department of Computer Science, Friedrich-Alexander-Universität Erlangen-Nürnberg (FAU), Germany}
\texttt{\small{\{muhammad.sabih, abrarul.karim, jakob.wittmann, frank.hannig, juergen.teich\}@fau.de}}%
}
\IEEEoverridecommandlockouts

\maketitle
\CopyrightNotice{-60,170}{-2.2ex}

\begin{abstract}
The customizability of RISC-V makes it an attractive choice for accelerating deep neural networks (DNNs). It can be achieved through instruction set extensions and corresponding custom functional units. Yet, efficiently exploiting these opportunities requires a hardware/software co-design approach in which the DNN model, software, and hardware are designed together. In this paper, we propose novel RISC-V extensions for accelerating DNN models containing semi-structured and unstructured sparsity. While the idea of accelerating structured and unstructured pruning is not new, our novel design offers various advantages over other designs. To exploit semi-structured sparsity, we take advantage of the fine-grained (bit-level) configurability of FPGAs and suggest reserving a few bits in a block of DNN weights to encode the information about sparsity in the succeeding blocks. The proposed custom functional unit utilizes this information to skip computations. To exploit unstructured sparsity, we propose a variable cycle sequential multiply-and-accumulate unit that performs only as many multiplications as the non-zero weights.
Our implementation of unstructured and  semi-structured pruning accelerators can provide speedups of up to a factor of 3 and 4, respectively.
We then propose a combined design that can accelerate both types of sparsities, providing speedups of up to a factor of 5.
Our designs consume a small amount of additional FPGA resources such that the resulting co-designs enable the acceleration of DNNs even on small FPGAs. We benchmark our designs on standard TinyML applications such as keyword spotting, image classification, and person detection.
\end{abstract}

\input{sec/sec_1}

\input{sec/sec_2}
\input{sec/sec_3}

\input{sec/sec_4}

\input{sec/sec_6}

\section*{Acknowledgment}
This work was partly supported by the Fraunhofer Institute for Integrated Circuits IIS, Erlangen, Germany, the Deutsche Forschungsgemeinschaft (DFG, German Research Foundation) under project number 524986327 (NA$^3$Os), and the Federal Ministry for Education and Research (BMBF) within project \enquote{DI-EDAI} (16ME0992).

%
%
\FloatBarrier


\printbibliography

\end{document}

%% file: head/header.tex
\IEEEoverridecommandlockouts

\usepackage{array}
\usepackage{float}
\usepackage{graphics, tikz, pgfplots}
\usepackage{lineno}
\usepackage{eso-pic}
\usepackage{epsfig}
\usepackage{amsmath}
\usepackage{caption}
\usepackage{multicol}
\usepackage{graphicx}

\usepackage[scaled=.95]{zi4} 

\usepackage{pifont}%
\newcommand{\cmark}{\ding{51}}%
\newcommand{\xmark}{\ding{55}}%

\usepackage[shrink=25]{microtype}
\usepackage{subcaption}
\usepackage{booktabs} 
\usepackage{scalefnt}
\usepackage[inline]{enumitem}
\usepackage{placeins}
\usepackage{multirow}
\usepackage{listings}
\usepackage{epsfig}
\usepackage{standalone}
\usepackage[nolist]{acronym}
\usepackage{algorithm}
\usepackage{algpseudocode}
\usepackage{xcolor}

\usepackage{pgf}
\usepackage{hyperref}
\usepackage{pgfplots}
\usepackage{pgfplotstable}
\usepackage{url}
\usepackage{todonotes}
\usepackage[autostyle]{csquotes}
\usepackage{lipsum}
\usetikzlibrary{
    shapes, positioning, calc, backgrounds,
    shapes.arrows, arrows,
    arrows.meta,
    decorations.pathmorphing,
    decorations.markings,
}
\usepgfplotslibrary{groupplots}
\usepackage{tikzscale}

\usepackage{amsfonts}       
\usepackage{nicefrac}       
\usepackage{environ}

\usepackage[backend=biber, style=numeric, hyperref, maxnames=8, minnames=1,
giveninits=true, maxcitenames=2, mincitenames=1, bibencoding=utf8, isbn=true,
sorting=none]{biblatex}

\definecolor{darkgreen}{rgb}{0.0, 0.5, 0.0}

\usepackage{cleveref}

\crefname{func}{Function}{Functions}
\Crefname{func}{Function}{Functions}
\definecolor{keywordcolor}{rgb}{0.26, 0.1, 0.75}
\definecolor{commentcolor}{rgb}{0.0, 0.5, 0.0}
\definecolor{stringcolor}{rgb}{0.58, 0.0, 0.0}

%% file: acronyms.tex
\newacro{AI}{artificial intelligence}
\newacro{AST}{abstract syntax tree}
\newacro{BCH}{Bose–Chaudhuri–Hocquenghem}
\newacro{BER}{bit error rate}
\newacro{CAM}{class activation map}
\newacro{CFU}{custom functional unit}
\newacro{CNN}{convolutional neural network}
\newacro{CV}{computer vision}
\newacro{DeepLIFT}{Deep Learning Important FeaTures}
\newacro{DNN}{Deep Neural Network}
\newacro{ECC}{error correcting code}
\newacro{FSM}{finite state machine}
\newacro{FPGA}{field Programmable gate array}

\newacro{FLOP}{floating-point operation}
\newacro{FLOPS}{floating-point operations per second}
\newacro{GPU}{graphics processing unit}
\newacro{IID}{independent and identically distributed}
\newacro{ILSVRC2012}{ImageNet Large Scale Visual Recognition Challenge~2012}
\newacro{ISA}{instruction set architecture}
\newacro{IWMSE}{importance-weighted mean square error}
\newacro{LRP}{layer-wise relevance propagation}
\newacro{LSB}{least significant bit}
\newacro{MAC}{multiply-accumulate operation}
\newacro{ML}{Machine Learning}
\newacro{MSB}{most significant bit}
\newacro{MSE}{mean square error}
\newacro{MOSP}{multi-objective sensitivity pruning}
\newacro{NAS}{neural architecture search}
\newacro{NP}{number of parameter}
\newacro{RISC}{reduced instruction set computer}
\newacro{RL}{reinforcement learning}
\newacro{SDFG}{synchronous data flow graph}
\newacro{SEU}{single-event upset}
\newacro{SHAP}{SHapley Additive exPlanations}
\newacro{SVD}{singular value decomposition}
\newacro{TMR}{triple modular redundancy}
\newacro{WCACFU}{weight clustering accelerator custom functional unit}
\newacro{XAI}{explainable AI}

%% file: sec/sec_1.tex
\section{Introduction}\label{sec:intro}
\Acp{DNN} are known to be computationally demanding, requiring significant amounts of computational resources and memory for both training and inference. \acp{DNN} are widely used in resource-constrained applications such as the \emph{Internet of Things} (IoT)~\cite{IoT}, \emph{EdgeAI}~\cite{EdgeAI}, \emph{TinyML}~\cite{TinyML}, etc. On the hardware side, advances in the semiconductor industry have made the development of \emph{custom AI accelerators} feasible.
RISC-V~\cite{RISCV_Manual} is an open standard \ac{ISA} that is quickly gaining traction in academia and industry. RISC-V has various customization possibilities, one of which is to design custom accelerator logic that is tightly coupled with the CPU and is readily utilized on the software side. This enables a smooth hardware/software co-design approach~\cite{HSCDTeich} that can unlock significant degrees of optimization potential.

Building ASICs is time-consuming and costly, and they are inflexible in terms of the \emph{flexibility-performance} tradeoff. On the other hand, general-purpose processors are not optimized for specialized DNN applications. A completely custom System-on-Chip (SoC) on an FPGA requires a significant development effort. RISC-V ISA extensions~\cite{RISCV_Manual} in the form of \acp{CFU}~\cite{CFU-Playground} offer an attractive tradeoff. Here, an SoC consisting of a general-purpose soft RISC-V core is used in combination with instruction extensions designed as hardware accelerators with custom logic that do not require a significant design effort but, at the same time, provide acceleration for DNN workloads~\cite{CFU-Playground, DATE_Sabih}.

\Acp{DNN} are typically over-parameterized; this means that pruning a DNN can often be carried out significantly without impacting its performance. The pruning of \acp{DNN} has been the subject of significant prior research~\cite{PruningSurvey, HSMSTH24}.
Pruning a neural network can involve removing neurons, filters, weights, or layers, typically leading to sparser DNNs.
The main goals are decreasing memory utilization, latency, and energy consumption~\cite{StreubuhrRHHT11}.
While DNN pruning approaches can be classified from various perspectives, one classification is based on \emph{structure}.
Based on the structure of sparsity, the DNN pruning method can be fully structured, semi-structured, or unstructured.
The different types of sparsity in a DNN model resulting from the different pruning methods are illustrated in \Cref{fig:typesOfSparsity}.

\input{figures/typesOfSparsity}

Fully structured pruning, such as layer or filter pruning (in the case of CNNs), is often accompanied by accuracy degradation. In comparison, the accuracy degradation for unstructured or semi-structured pruning is significantly less. Many popular DNN architectures have been pruned using unstructured pruning with high sparsity ratios~\cite{SOTA_Pruning}.
A \emph{sparsity ratio} $x$ is defined as the percentage of zeros in a model.
Unstructured or semi-structured pruning has a drawback; obtaining acceleration from it on general-purpose processors is infeasible.

\begin{figure*}[t]
\centering
    \resizebox{0.9\textwidth}{!}{
        \input{figures/VA}
    }
\caption{Overview of our method for hardware/software co-design of RISC-V extensions to accelerate sparse \acp{DNN} on FPGAs.
The process starts with a DNN model, which is pruned using unstructured or semi-structured pruning. Software customization of DNN kernels and hardware specialization of RISC-V extensions are jointly performed according to the co-design approach.}
\label{fig::diagram}
\end{figure*}

CPU-based architectures most often only support the execution of dense computations and uniform data structures and perform the processing of sparse computations much less efficiently. One approach to performing sparse computations is to store sparse matrices in compressed format, which retains non-zero elements and indexing information~\cite{CSR}. This approach is useful for very high sparsity (90\%--99\%).
For \acp{DNN}, the sparsity is high but not nearly as high as this.
Therefore, the compressed storage approach becomes unsuitable.
This strongly motivates exploiting customizable hardware to accelerate modern \acp{DNN} pruned with semi-structured or unstructured pruning.

Literature on accelerating sparse \acp{DNN} (unstructured or semi-structured) can be classified into two categories.
The first category targets general-purpose processors, including CPUs and GPUs.
For example, NVIDIA introduced a 2:4 pruning scheme that can be accelerated on NVIDIA Ampere architecture GPUs~\cite{2:4Pruning}. In \cite{SparseRT}, the authors propose a code generator for accelerating sparse DNN models on GPUs.
In the second category of works, customized accelerator architectures are proposed for accelerating sparse \acp{DNN}, such as SNAP~\cite{SNAP} and DANNA~\cite{DANNA}. To the best of our knowledge, existing work on extending the instruction set architectures for accelerating sparse models is limited. In IndexMAC~\cite{IndexMAC}, the authors accelerated models with structured sparsity by extending a RISC-V CPU and demonstrated a speedup of 1.80--2.14$\times$.
Utilizing completely general-purpose processors is less efficient, while using solely custom accelerators is less flexible and not readily available.
The advantage of proposing instruction extensions of a RISC-V is that it offers a decent tradeoff between the two cases. In a hardware/software co-design approach, hardware customization, software specialization, and model (DNN) optimization are jointly approached. The result is a modest increase in hardware resources and substantial performance speedups over CPU-only implementations as will be demonstrated.

With this motivation, we propose accelerating \acp{DNN} by exploiting semi-structured and unstructured sparsity using a hardware/software co-design approach and implementing this approach by providing instruction set extension units in hardware for RISC-V CFUs (illustrated in Fig.~\ref{fig::diagram}). Our contributions are as follows:

\begin{enumerate}[leftmargin=1.2\parindent,itemsep=0cm]
    \item We propose instruction set extensions for RISC-V CPUs to support \emph{unstructured} sparsity for accelerated DNN processing at a modest increase in FPGA resource usage. These can be used for both TinyML and normal DNN workloads. In contrast to other approaches, our design makes no assumptions on the structure and number of zeros.

    \item Instruction set extensions for RISC-V CPUs are proposed to support \emph{semi-structured} sparsity with accompanied software specialization. A novel lookahead encoding scheme is proposed here that does not compromise the DNN's performance.

    \item We introduce a combined design for accelerating a DNN to support both unstructured pruning and semi-structured pruning. This dual-pruning capability is beneficial because it allows the model to simultaneously leverage each pruning method's distinct degrees of freedom, thereby enhancing computational efficiency and reducing model complexity.
\end{enumerate}

%% file: figures/typesOfSparsity.tex
\begin{figure}[t]
\hspace{1mm}
\begin{subfigure}[b]{28mm}
    \begin{tikzpicture}[scale=0.3]
      \fill[blue!25] (5,8) rectangle (4,0);
      \fill[blue!25] (8,6) rectangle (0,5);
      \draw[step=1cm,black,very thin] (0,0) grid (8,8);
    \end{tikzpicture}\vspace{-1mm}
    \caption{\ \ \ }
  \end{subfigure}
  \hfill
  \begin{subfigure}[b]{28mm}
    \begin{tikzpicture}[scale=0.3]
      \fill[blue!25] (1,5) rectangle (0,1);
      \fill[blue!25] (2,3) rectangle (1,2);
      \fill[blue!25] (1,7) rectangle (0,6);
      \fill[blue!25] (2,6) rectangle (1,5);
      \fill[blue!25] (4,7) rectangle (2,6);
      \fill[blue!25] (5,8) rectangle (1,7);
      \fill[blue!25] (8,8) rectangle (6,7);
      \fill[blue!25] (8,7) rectangle (7,5);
      \fill[blue!25] (5,5) rectangle (2,2);
      \fill[blue!25] (6,6) rectangle (5,3);
      \fill[blue!25] (6,2) rectangle (4,0);
      \fill[blue!25] (8,1) rectangle (2,0);
      \fill[blue!25] (8,3) rectangle (6,2);
      \fill[blue!25] (7,5) rectangle (6,4);
      \draw[step=1cm,black,very thin] (0,0) grid (8,8);
    \end{tikzpicture}\vspace{-1mm}
    \caption{\ \ \ }
  \end{subfigure}
  \hfill
  \begin{subfigure}[b]{28mm}
    \begin{tikzpicture}[scale=0.3]
      \fill[blue!25] (3,8) rectangle (1,7);
      \fill[blue!25] (8,8) rectangle (6,7);

      \fill[blue!25] (4,7) rectangle (2,6);
      \fill[blue!25] (6,7) rectangle (5,6);
      \fill[blue!25] (8,7) rectangle (7,6);

      \fill[blue!25] (2,6) rectangle (0,5);
      \fill[blue!25] (7,6) rectangle (5,5);

      \fill[blue!25] (1,5) rectangle (0,4);
      \fill[blue!25] (3,5) rectangle (2,4);
      \fill[blue!25] (6,5) rectangle (4,4);

      \fill[blue!25] (2,4) rectangle (1,3);
      \fill[blue!25] (4,4) rectangle (3,3);
      \fill[blue!25] (6,4) rectangle (4,3);

      \fill[blue!25] (3,3) rectangle (1,2);
      \fill[blue!25] (6,3) rectangle (5,2);
      \fill[blue!25] (8,3) rectangle (7,2);

      \fill[blue!25] (4,2) rectangle (2,1);
      \fill[blue!25] (5,2) rectangle (4,1);
      \fill[blue!25] (7,2) rectangle (6,1);

      \fill[blue!25] (2,1) rectangle (1,0);
      \fill[blue!25] (4,1) rectangle (3,0);
      \fill[blue!25] (6,1) rectangle (5,0);
      \fill[blue!25] (8,1) rectangle (7,0);

      \draw[step=1cm,black,very thin] (0,0) grid (8,8);
      \draw[xstep=4cm,ystep=1cm,orange,thick] (0,0) grid (8,8);
    \end{tikzpicture}\vspace{-1mm}
    \caption{\ \ \ }
  \end{subfigure}
  \caption{Different sparsity structures: (a) Structured sparsity, resulting from structured pruning, which removes whole columns or rows from, e.g., a convolution matrix. (b) Unstructured sparsity, resulting from unstructured pruning, which removes arbitrary weights. (c) Semi-structured sparsity, resutling from semi-structured pruning, a.k.a.~$n$:$m$ pruning that zero-outs $n$ weights every $m$ elements; shown is a 2:4 pattern.}\label{fig:typesOfSparsity}\vspace{-3mm}
\end{figure}

%% file: figures/VA.tex
\begin{tikzpicture}[node distance=2cm, auto]

\def\minwidth{4.7cm}
\def\minheight{3.8cm}

\def\textfontsize{\Large} 

\node [draw, rectangle, very thick, text width=2.2cm, minimum height=2.4cm, align=center] (model) {
    \ifuseicons
        \includegraphics[width=1cm]{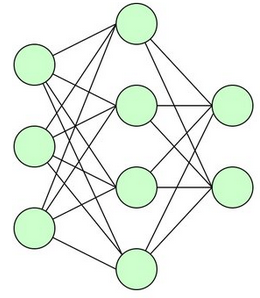} \\
    \fi
    {\textfontsize DNN}
};
\node [draw, rectangle, very thick, text width=\minwidth, minimum height=3.8cm, align=center, right=of model] (pruning) {
    \ifuseicons
        \includegraphics[width=1cm]{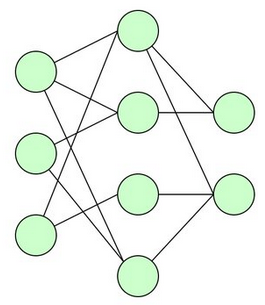} \\
    \fi
    {\textfontsize DNN Model Pruning\\(Unstructured/Semi-Structured)}
};
\node [draw, rectangle, very thick, text width=\minwidth, minimum height=3.8cm, align=center, right=of pruning] (customization) {
    \ifuseicons
        \includegraphics[width=1cm]{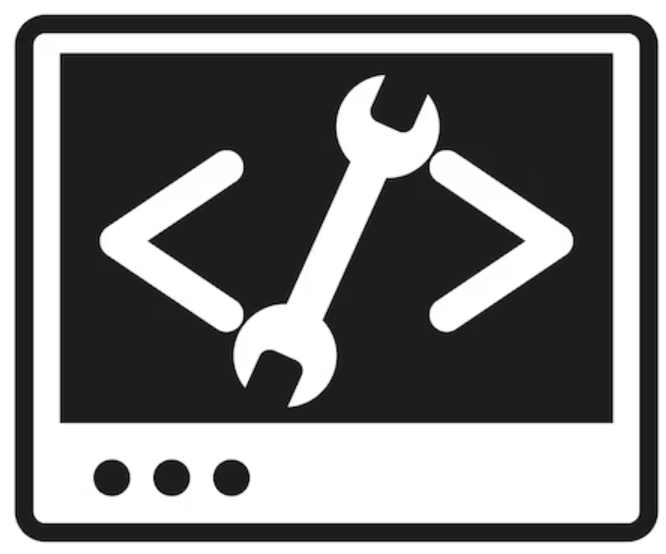} \\
    \fi
    {\textfontsize Software \vspace{-0.15em}\\ Customization \\ \vspace{0.3em} (DNN kernels)}
};
\node [draw, rectangle, very thick, text width=\minwidth, minimum height=\minheight, align=center, right=of customization] (specialization) {
    \ifuseicons
        \includegraphics[width=1cm]{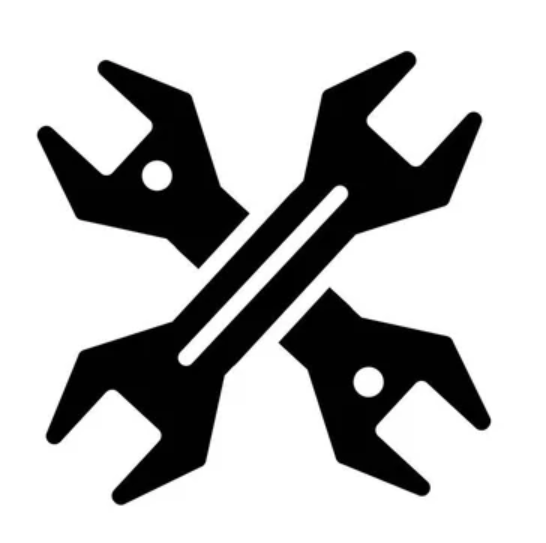} \\
    \fi
    {\textfontsize Hardware \vspace{-0.15em} \\Specialization\\(RISC-V Extensions)}
};
\node [draw, rectangle, very thick, text width=2.2cm, minimum height=2.4cm, align=center, right=of specialization] (fpga) {
    \ifuseicons
        \includegraphics[width=1cm]{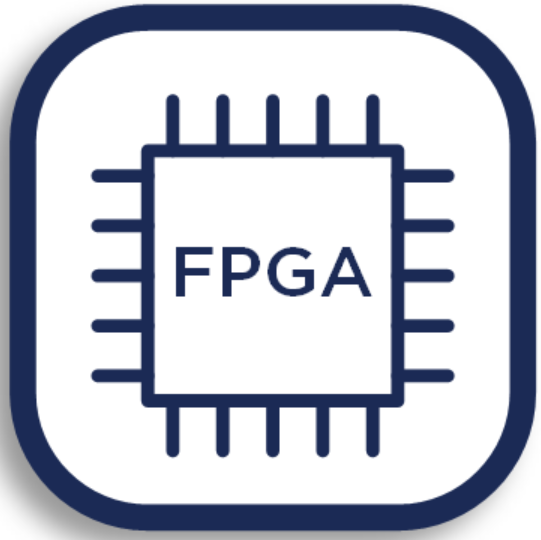} \\
    \fi
    {\textfontsize \vspace{0.30em} FPGA Target}
};

\node [fit=(customization)(specialization), inner sep=0.9cm, draw=cyan, very thick, rounded corners, dashed] (cfu-box) {};
\node [above=0.00cm of cfu-box.south] (cfu-text) {\textfontsize Hardware/Software Co-Design};

\draw [->, thick, gray, line width=1mm] (model) -- (pruning);
\draw [->, thick, gray, line width=1mm] (pruning) -- (customization);
\draw [->, thick, gray, line width=1mm] (specialization) -- (fpga);
\draw [<->, thick, gray, line width=1mm] (customization) -- (specialization);

\end{tikzpicture}

%% file: sec/sec_2.tex
\section{Background and Preliminaries}
This section provides a brief overview of the RISC-V \acf{ISA}~\cite{RISCV_Manual} and CFU~Playground~\cite{CFU-Playground}.
\subsection{Custom Functional Units}
\begin{figure}[b]
    \centering
    \vspace{-1em}
    \includegraphics[width=1.1\linewidth]{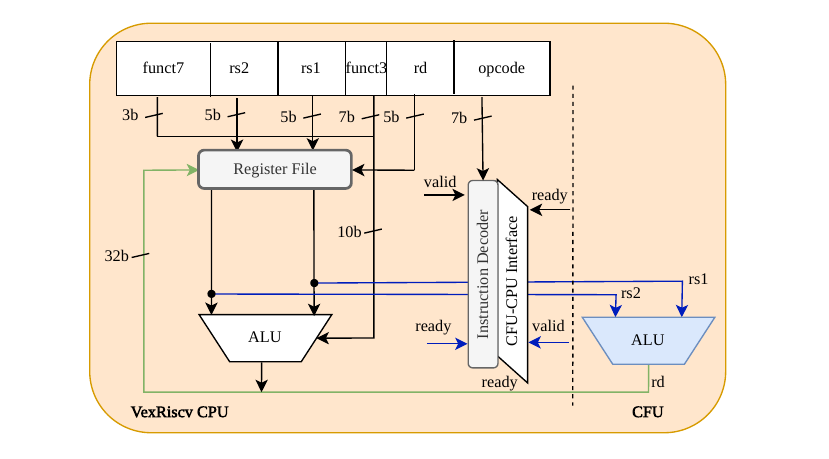}
    \caption{CPU-CFU interface using R-type instruction of RISC-V.}
    \label{fig:Rtype}
\end{figure}

The RISC-V \ac{ISA}~\cite{RISCV_Manual} allows for tailored instruction set extensions and accelerator design.
Instruction encoding spaces and variable-length encoding make this accessible, letting developers customize processors while still using the standard ISA toolchain.
Customization of the RISC-V processor architecture is achieved with so-called \acp{CFU}.
These \acp{CFU} refer to custom logic added in hardware that is tightly coupled with the processor to provide specialized functionality.
A \ac{CFU} is addressed by the RISC-V ISA using the \emph{R-type} instruction.
The format of an R-type 32-bit instruction is shown in \Cref{fig:Rtype} and can be described as follows:
The \emph{opcode} (7 bits) together with 3-bit \emph{funct3} and 7-bit \emph{funct7} specify the type of the instruction format and the operation to be performed.
Fields \emph{rs1} and \emph{rs2} denote two source registers and \emph{rd} the destination register, each addressed by 5~bits.

Notably, CFUs do not have direct access to the main memory.
Therefore, the CPU acts as an intermediate node to transfer the data in memory to the \ac{CFU}.
Handshaking between CPU and \ac{CFU} is handled using valid and ready signals. General pipeline stages can be described as follows. When the field opcode matches \emph{custom-0} (predefined for custom instructions \cite{RISCV_Manual}), the CPU recognizes that the instruction needs to be forwarded to the CFU.
At the same time, the register file resolves the \emph{rs1} and \emph{rs2} addresses to two 32-bit values.
These bits form the input to the \ac{CFU} along with the \emph{funct7} and \emph{funct3} values.
Once a computation has finished inside the \ac{CFU}, which can take one or multiple clock cycles, a valid signal is set to notify the CPU, and the
result is written back from the \ac{CFU} to the register file.

\subsection{CFU Playground}
Building custom instructions can be a tedious process as it requires hardware/software co-design along with the ability to profile, modify, and rapidly prototype. With the goal of enabling a rapid exploration and design of \acp{CFU}, CFU Playground~\cite{CFU-Playground} was developed. The CFU Playground workflow can be divided into three main stages: \emph{deployment}, \emph{profiling}, and \emph{optimization}. A TensorFlow Lite model is given to CFU Playground, which uses a RISC-V compiler alongside SymbiFlow\footnote{Open-source flow for generating bitstreams from Verilog (\url{https://github.com/SymbiFlow})} or Vivado for synthesis in the \emph{deployment} stage. Then, in the \emph{profiling} stage, various options are available to identify suitable candidates for optimization using custom functional units. Next, in the \emph{optimization} phase, custom instructions are designed, and the three stages are repeated. CFU Playground uses an open-source implementation of a RISC-V processor known as VexRiscv~\cite{VexRiscv}. A LiteX SoC configuration is placed within VexRiscv. The CPU-CFU interface provides a very tight coupling between the CPU and the added custom functionality. During FPGA synthesis, place, and route, the interface disappears, and the CFU essentially becomes a CPU pipeline component. The new instructions can be utilized in C or C++ application programs through an inline assembly macro provided. No adjustments to the RISC-V GCC toolchain are necessary.

%% file: sec/sec_3.tex
\section{Proposed approach}
This section provides a comprehensive breakdown of our proposed approach. First, we describe the basics (\Cref{ssec::baseline}), followed by proposing hardware designs to support semi-structured (\Cref{ssec:SSSA}), unstructured pruning (\Cref{ssec:USSA}) and finally the combined design (\Cref{ssec:CSA}).

\subsection{Baseline}\label{ssec::baseline}
Our starting point is a VexRiscv soft-core with five pipeline stages and a CFU implementing a Single Instruction, Multiple Data (SIMD) MAC instruction (\texttt{cfu\_simd\_mac}) that takes four INT8 weights (\texttt{filter[i]}) and four INT8 activations/inputs (\texttt{input[i]}) as two inputs of a custom instruction and returns the multiply-and-accumulate result. This initial design is provided by TFLite\footnote{TensorFlow Lite (TFLite) is a collection of tools to convert and optimize TensorFlow~\cite{tensorflow} models to run on mobile and edge devices} included within CFU Playground.

\Cref{lst::unoptimized_conv} displays the baseline pseudo-code of a convolutional kernel with kernel height \texttt{output\_height}, kernel width \texttt{output\_width}, and output channels \texttt{output\_channels}, which utilizes this baseline CFU design as a SIMD MAC instruction (\texttt{cfu\_simd\_mac}).

\lstset{
    language=C++,
    frame=single,
    breaklines=true,
    basicstyle=\scriptsize\ttfamily,
    keywordstyle=\color{keywordcolor},
    commentstyle=\color{darkgreen},
    stringstyle=\color{stringcolor},
}

\begin{lstlisting}[language=C,frame=single,breaklines=true,
    basicstyle=\footnotesize\ttfamily, label=lst::unoptimized_conv, caption=Pseudo-code of baseline convolutional kernel.]
for (output_height) {
  for (output_width) {
    for (out_channel) {
      for (int i=0; i<in_channel; i+=4) { // 4x4 MAC
        cfu_simd_mac(filter[i], input[i]); }}}}
\end{lstlisting}

\subsection{Semi-Structured Sparsity Accelerator (SSSA)}\label{ssec:SSSA}
In the baseline design (see \Cref{lst::unoptimized_conv}), a single call to \texttt{cfu\_mac} multiplies four INT8 values and returns the accumulated sum. In semi-structured pruning, sparsity manifests as blocks of zero weights.
Therefore, to efficiently exploit sparsity, blocks of consecutive zeros must be skipped.
One possible approach is to design a CFU that only processes non-zero blocks of weights and simply skips all other blocks; however, this introduces overhead within the innermost loop.

Our approach involves co-designing hardware and software components tailored to the DNN application.
Notably, DNN weights remain static at runtime, and FPGAs provide bit-level granularity. Exploiting these two characteristics, we propose a pre-processing step of the list of weights for calculating the number of consecutive all-zero blocks following each non-zero block and then encoding this number into the non-zero block weights (see \Cref{alg:encoding}). At execution time, this encoded counter is extracted in hardware and used for incrementing an induction variable in the innermost loop through custom instructions (\Cref{lst::opt_struc}).

\begin{algorithm}[b]
\caption{Encode CNN Kernel Weights with Lookahead Information}
\label{alg:encoding}
\begin{algorithmic}[1]
\footnotesize
\Require CNN kernel represented as a 3D matrix (\texttt{kernel})
\Ensure Encoded CNN kernel with lookahead information
\State Initialize kernel dimensions: \texttt{C} (number of input channels), \texttt{H} (height),  \texttt{W}~(width)
\For{\texttt{h} = 0 to \texttt{H}-1}
    \For{\texttt{w} = 0 to \texttt{W}-1}
        \For{\texttt{c} = 0 to \texttt{C}-1 step 4}
            \State \texttt{i\_nxt} $\gets$ \texttt{c} + 4
            \State \texttt{skip\_blocks} $\gets$ 0
            \While{\texttt{i\_nxt} \textless\ \texttt{C} and \texttt{skip\_blocks} \textless\ 4}
                \If{\texttt{checkBlkSkip}(\texttt{kernel[h][w][i\_nxt]})}
                    \State \texttt{skip\_blocks} $\gets$ \texttt{skip\_blocks} + 1
                    \State \texttt{i\_nxt} $\gets$ \texttt{i\_nxt} + 4
                \Else
                    \State \textbf{break}
                \EndIf
            \EndWhile
            \For{\texttt{k} = 0 to 3}
                \If{\texttt{c} + \texttt{k} \textless\ \texttt{C}}
                    \State \texttt{encodeLastBits}(\texttt{kernel[h][w][c+k],skip\_blocks})
                \EndIf
            \EndFor
        \EndFor
    \EndFor
\EndFor
\State \Return Encoded CNN kernel
\end{algorithmic}
\end{algorithm}

\begin{figure*}[t]
    \centering
    \includegraphics[width=0.83\textwidth]{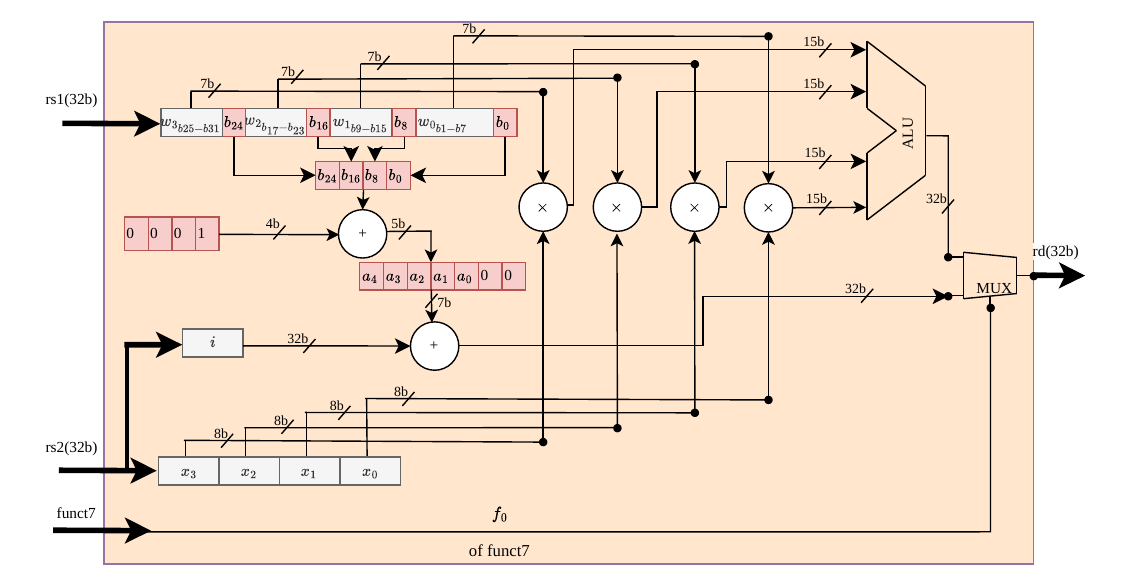}
    \caption{RTL diagram of the proposed hardware SSSA for exploiting semi-structured sparsity.}
    \label{fig::SSSA_RTL}
\end{figure*}

\Cref{alg:encoding} processes a given 3D matrix of CNN weights (\texttt{kernel}) with dimensions corresponding to the number of input channels ($C$), height ($H$), and width ($W$). The algorithm iterates over each kernel element in the height and width dimensions, testing four consecutive blocks of weights along the input channel dimension for being zero or not.

\begin{figure}
\hspace{-5.5mm}\includegraphics[width=1.1\linewidth]{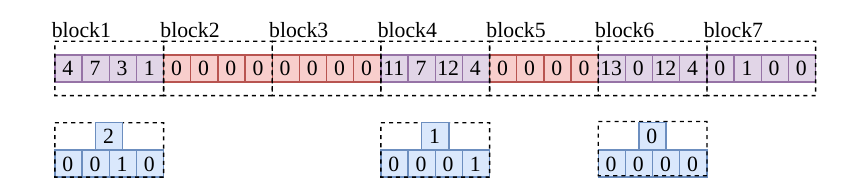}
    \caption{The first row shows 7 blocks of DNN weights, each containing four INT8 weights. \Cref{alg:encoding} (encoding) performs a pass over all blocks, annotates each non-zero block with the information of the number of succeeding all-zero blocks, and encodes their number. The 2nd row shows the calculated code.}
    \label{fig:enc1}
\end{figure}

When a block of four consecutive zeros is detected, the \texttt{skip\_blocks} counter is incremented, which can range from 0 to 15. This counter indicates how many all-zero blocks can be skipped during computation, thus optimizing the execution by reducing unnecessary operations. An example of this step is illustrated in \Cref{fig:enc1}.

\begin{figure}
\hspace{-3.5mm}\includegraphics[width=1.08\linewidth]{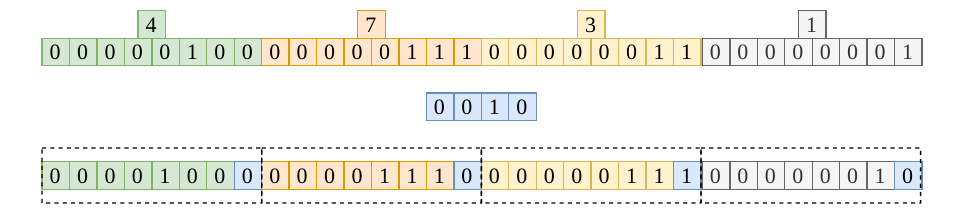}
    \caption{The first row shows four weights along with their binary representations, and the second row shows the corresponding 4-bit code obtained in \Cref{fig:enc1}. The sign bit of the weight is saved, and each bit in the code is appended to the LSB of the corresponding weight.}
    \label{fig:enc2}
\end{figure}

The dynamic range of INT8 weights is limited to [-64, 63] so as to not use the most significant bit after the signed bit, effectively simulating INT7 precision. The EncodeLastBits function (see \Cref{alg:encodeLastBits}) embeds the 4-bit skip\_blocks value into each block of four DNN weights by shifting the bits of each weight to the left, making space for the skip bit, and then inserting the skip bit into the least significant bit (LSB) of each weight.
The encoding procedure is visualized in \Cref{fig:enc2}.
By encoding the skip information directly within the weights, our hardware design, as shown in \Cref{fig::SSSA_RTL}, can directly use the extracted information of as many blocks to skip to increment an induction variable (\texttt{i}) in the innermost loop, as shown in \Cref{lst::opt_struc}. Thereby, we can avoid any runtime overhead in software tests.

According to Line 12 of \Cref{alg:encodeLastBits}, the LSB of a given weight is not overwritten directly but appended as the LSB with all higher significant bits shifted left. Our experiments (see \Cref{{ssec:sacrifice}}) show that \enquote{sacrificing} one bit does not adversely affect the DNN accuracy for the considered applications. Next, we describe 1) the details of the hardware design and 2) the kernel customization more closely.

\begin{algorithm}
\caption{encodeLastBits}
\label{alg:encodeLastBits}
\begin{algorithmic}[1]
\footnotesize
\Function{encodeLastBits}{weights[4], skip\_blocks}
    \For{\texttt{i} = 0 to 3}
        \State \textcolor{darkgreen}{/* Isolate the sign bit */}
        \State \texttt{sign\_bit} $\gets$ (\texttt{weights[i]} \textgreater\textgreater\ 7) \& 0b1
        \State \textcolor{darkgreen}{/* Extract skip bit */}
        \State \texttt{skip\_bit} $\gets$ (\texttt{skip\_blocks} \textgreater\textgreater\ \texttt{i}) \& 0b1
        \State \textcolor{darkgreen}{/* Remove the MSB after the sign bit */}
        \State \texttt{weights[i]} $\gets$ \texttt{weights[i]} \& 0b10111111
        \State \textcolor{darkgreen}{/* Shift bits one position to the left */}
        \State \texttt{weights[i]} $\gets$ (\texttt{weights[i]} \textless\textless\ 1) \& 0b01111110
        \State \textcolor{darkgreen}{/* Insert skip bit */}
        \State \texttt{weights[i]} $\gets$ \texttt{weights[i]} \textbar\ \texttt{skip\_bit}
        \State \textcolor{darkgreen}{/* Restore the sign bit */}
        \State \texttt{weights[i]} $\gets$ \texttt{weights[i]} \textbar\ (\texttt{sign\_bit} \textless\textless\ 7)
    \EndFor
    \State \Return \texttt{weights}
\EndFunction
\end{algorithmic}
\end{algorithm}

\begin{figure*}[!h]
    \centering
    \includegraphics[width=0.83\textwidth]{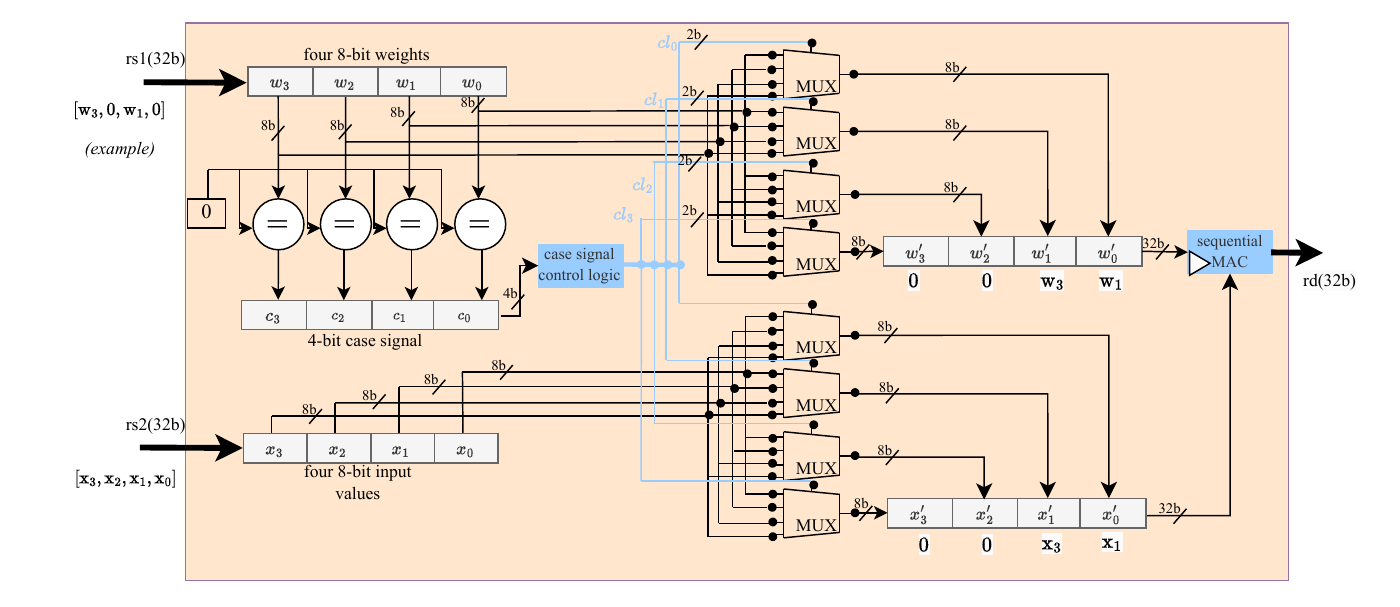}
    \caption{RTL diagram of the proposed USSA for exploiting unstructured sparsity.}
    \label{fig:rtl_ussa}
\end{figure*}

\subsubsection{Hardware Design}\label{SSSA:hardware_design}
The CFU, as shown in \Cref{fig::SSSA_RTL}, implements two instructions (\texttt{sssa\_mac} and \texttt{sssa\_inc\_indvar}): \texttt{sssa\_inc\_indvar} takes the LSB (Least Significant Bit) of each of the four weights (first operand of the instruction) to offset the induction variable (\texttt{i}), while \texttt{sssa\_mac} performs a MAC operation on four 7-bit weights and four 8-bit input values.
The LSB of the \texttt{funct7} field is used to choose between these two instructions.

In \Cref{fig::SSSA_RTL}, the first operand, given by \texttt{$rs1$}, a 32-bit input register, provides four consecutive 7-bit weights \((w_3, w_2, w_1, w_0)\) augmented by the corresponding encoding information bits \((b_{24}, b_{16}, b_8, b_0)\). The latter are extracted to form a 7-bit increment value. The second operand, given by \texttt{$rs2$}, a 32-bit input register, is either interpreted as the value of the induction variable \texttt{$i$} or four 8-bit input values \((x_3, x_2, x_1, x_0)\) depending on the particular instruction used, which is differentiated by the LSB of \texttt{funct7} ($f_0$). If \texttt{funct7} indicates a MAC operation, four 7-bit weights and four 8-bit values are multiplied and accumulated. Otherwise, the lookahead information contained in \((b_{24}, b_{16}, b_8, b_0)\) of \texttt{$rs1$} is used to increment the induction variable \texttt{$i$} by multiples of four. This is achieved by adding one to the bits encoding skip blocks information \((b_{24}, b_{16}, b_8, b_0)\) and left shifting by two to multiply by four, obtaining the eventual 7-bit increment \((a_{4}, a_{3}, a_{2}, a_1, a_0, 0, 0)\), which leads to skipping the specified number of zero-blocks in the innermost loop, as summarized in \Cref{lst::opt_struc}.

\lstset{
    language=C++,
    frame=single,
    breaklines=true,
    basicstyle=\scriptsize\ttfamily,
    keywordstyle=\color{keywordcolor},
    commentstyle=\color{darkgreen},
    stringstyle=\color{stringcolor},
}

\begin{lstlisting}[language=C,frame=single,breaklines=true,
    basicstyle=\footnotesize\ttfamily, caption={Pseudo-code of specialized kernel for SSSA.}, label=lst::opt_struc]
for (output_height){
  for (output_width){
    for (out_channel){
      int i = 0;
      while (i<in_channel){
        sssa_mac(filter[i], input[i]); // 4x4 MAC
        i = sssa_inc_indvar(filter[i], i);}}}}
\end{lstlisting}

\subsubsection{Kernel customization}
In order to use the design unit, the kernel shown in \Cref{lst::unoptimized_conv} is modified to \Cref{lst::opt_struc} as follows: we replace the \texttt{for} loop with a \texttt{while} loop in the innermost loop of the baseline code for a convolutional kernel. Our two custom instructions are then inserted in this \texttt{while} loop.
Instruction \texttt{sssa\_inc\_indvar} processes four weights packed as one 32-bit operand (including four bits for lookahead information) \texttt{filter[i]} and the current value of the induction variable (\texttt{i}), returning the updated value of the induction variable.
Based on the encoded lookahead information in the block of weights, up to a maximum of 15 subsequent blocks can be skipped.
The other instruction, \texttt{sssa\_mac}, multiplies four 7-bit weights with four 8-bit input values as described in \Cref{SSSA:hardware_design}.

%% file: sec/sec_4.tex
\subsection{Unstructured Sparsity Accelerator (USSA)}\label{ssec:USSA}

While semi-structured sparsity will be shown to offer a notable acceleration, it still has the finest granularity of only full blocks and limits the number of blocks to be skipped. In the following, we suggest a co-design solution for fully unstructured pruning. The corresponding design can be combined and integrated with previous approaches to accelerate both unstructured and semi-structured pruning. The following sections outline our proposed approach for supporting the acceleration of unstructured sparsity.

\subsubsection{Baseline}
Our approach employs a baseline \emph{single} sequential MAC unit that multiplies four input values by four weight values over \emph{four} cycles, returning the accumulated sum. This baseline sequential design consistently requires \emph{four clock cycles} regardless of the presence of zeros in the weights.

\subsubsection{Hardware Design}

Proposed is a variable-cycle MAC unit that takes only as many cycles as non-zero weight elements in a block except a single cycle for an all-zero block. In our RISC-V design shown in \Cref{fig:rtl_ussa}, each block contains four INT8 weights.
The input to the CFU comes through two 32-bit input registers (shown on the left side in the figure); these are four weights \((w_3, w_2, w_1, w_0)\) and four input values \((x_3, x_2, x_1, x_0)\). Each of the four weights is compared in parallel to zero, generating 4-bit case signal \((c_3, c_2, c_1, c_0)\). The case signal control logic then processes this signal to produce four control signals \((cl_0, cl_1, cl_2, cl_3)\), which dictate the selection logic of the following two sets of multiplexers. The multiplexers, each controlled by the control signals, selectively pass and align the input weights \((w'_3, w'_2, w'_1, w'_0)\) and input values \((x'_3, x'_2, x'_1, x'_0)\). After the data is aligned, it is passed to a sequential MAC unit that takes as many cycles as non-zero weights in the block, with the exception of an all-zero block, in which case one cycle is taken.

\subsubsection{Kernel Customization}
Utilizing unstructured sparsity requires a simple modification: replacing the baseline MAC (see \Cref{lst::unoptimized_conv}) instruction with the \texttt{usss\_vcmac}, which exploits the unstructured sparsity via our variable-cycle sequential MAC design. The considered baseline performs the multiplications sequentially since we consider the case of a \emph{single} multiplier. Whereas the baseline MAC unit always takes \emph{four} clock cycles per block. In contrast, \texttt{usss\_vcmac} takes a \emph{variable} number of clock cycles depending on the number of zeros.

\subsection{Combined Sparsity Accelerator (CSA)}\label{ssec:CSA}
Finally, we propose the Combined Sparsity Accelerator (CSA), which integrates the functionalities of both prior designs. In practice, a combined approach using both semi-structured and unstructured sparsity
techniques maximizes performance gains because DNN models often exhibit both sparsity types and can be optimized for both types of sparsities. This dual capability allows for more degrees of freedom during pruning a DNN. The combined design has two instructions: \texttt{csa\_inc\_indvar} and \texttt{csa\_vcmac} (shown in \Cref{lst::opt_csa}).
\texttt{csa\_inc\_indvar} behaves in the same way as \texttt{sssa\_inc\_indvar} instruction introduced before our SSSA design to accelerate the semi-structured sparsity while \texttt{csa\_vcmac} is just a variable-cycle MAC unit similar to the \texttt{usss\_vcmac} in our USSA design to accelerate the unstructured pruning except that the weights are considered to be of 7 bits. The CSA combines the advantages of both the USSA and the SSSA.

\vspace{-1.5mm}
\lstset{
    language=C++,
    frame=single,
    breaklines=true,
    basicstyle=\scriptsize\ttfamily,
    keywordstyle=\color{keywordcolor},
    commentstyle=\color{darkgreen},
    stringstyle=\color{stringcolor},
}
\begin{lstlisting}[language=C++,frame=single,breaklines=true,
    basicstyle=\footnotesize\ttfamily,
label=lst::opt_csa,
 caption={Pseudo-code of the specialized kernel for CSA.}, label=lst::opt_csa]
for (output_height){
  for (output_width){
    for (out_channel){
      int i = 0;
      while (i<in_channel){
        csa_vcmac(filter[i], input[i]);
        // 4x4 variable-cycle sequentialMAC
        i = csa_inc_indvar(filter[i], i);}}}}
\end{lstlisting}

%% file: sec/sec_6.tex
\section{Experiments}\label{eval}
In this section, we assess our proposed designs in terms of achievable speedup and FPGA resource utilization.

\subsection{Evaluation Methodology}\label{eval:methodology}
Our designs were implemented on an Arty A7-35T FPGA board. Although our experiments primarily focus on convolutional and fully connected layers within CNNs, our designs can be seamlessly adapted to support other layer types as well, such as LSTMs and fully connected layers without any further instruction set extension or modification.

\subsection{Models and Datasets}
Three datasets and four models are used in our evaluation. Two models are used with CIFAR-10~\cite{CIFAR10}, which is a dataset with 50,000 training and 10,000 testing 32x32 color images divided into ten classes. Keyword Spotting focuses on recognizing specific words. The Google Speech Commands (GSC) v2 dataset~\cite{SpeechCommands} contains 65,000 audio signals spoken by a diverse group of individuals, and it consists of 30 audio classes such as \enquote{up,} \enquote{down,} \enquote{yes,} \enquote{no,} etc. Person detection refers to a machine learning task where the goal is to detect the presence or absence of a person in an image. The Visual Wake Words (VWW) dataset~\cite{VisualWake} was derived from the COCO dataset~\cite{COCO} and includes around 115,000 training and validation images. We evaluate four models: VGG16, ResNet-56, MobileNetV2, and DSCNN.

\subsection{Pruning Methodology}
We have not delved into training any pruned DNN model and optimizing for accuracy due to space constraints and because our primary focus is on proposing RISC-V extensions to accelerate structured and unstructured sparsity. In principle, any pruning method that generates a model as input with unstructured sparsity or semi-structured sparsity conforming to our sparsity pattern can be utilized.
For our purposes, we applied an iterative pruning approach with explainable-AI-based ranking similar to the methods described in \cite{SHT20}, \cite{SHT22a}, and \cite{sabih_mosp}.

\subsection{Speedup for USSA}\label{eval:ussa}
The USSA (Unstructured Sparsity Accelerator) is used to accelerate models with unstructured sparsity. No constraints are enforced on the structure of the sparsity. In the baseline scenario, no computational reductions are made to exploit the sparsity.

The impact of sparsity ($x \in [0,1]$) on speedup can be quantified in general by examining the probability distribution of zeros and ones within DNN weights assuming an IID\footnote{independent, identically distributed (IID)} distribution. In that case, the analytical average number of clock cycles ($c_a$) can be computed as:

\begin{align}
  c_a = \sum_{k=0}^{4} \binom{4}{k} \cdot x^k \cdot (1-x)^{4-k} \cdot (4-k)\nonumber
\end{align}

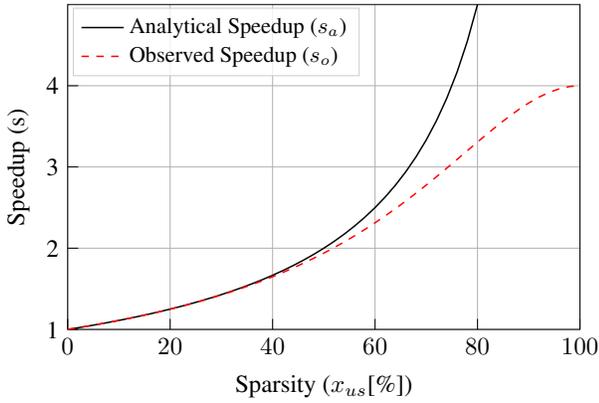
\begin{figure}[t]
\centering
\resizebox{0.45\textwidth}{!}{\input{figures/fig1_iccad}}
\caption{Analytical and observed speedups for USSA (Unstructured Sparsity Accelerator).}
\label{fig::speedup}
\end{figure}

In this equation, \(\binom{4}{k}\) is the binomial coefficient representing the number of ways to choose \(k\) ones in a block of four elements, and \(4-k\) denotes the number of cycles required for each configuration. In the ideal case, zero clock cycles are needed for a block of four zeros. In contrast, our design still requires one clock cycle for a block of four zeros. The observed average number of clock cycles ($c_o$) can then be given as:

\begin{align}
  c_o &= \sum_{k=0}^{3} \binom{4}{k} \cdot x^k \cdot (1-x)^{4-k} \cdot (4-k) + \binom{4}{4} \cdot x^4 \cdot (1-x)^0\nonumber
\end{align}
The analytical speedup ($s_a$) and observed speedup ($s_o$) can be obtained as $s_a = 4/c_a$ and $s_o = 4/c_o$, respectively. A comparison of the two speedups is shown in \Cref{fig::speedup}. The effect of a single cycle overhead when all four weights in a block are zero becomes only noticeable at very high sparsities. However, this additional cycle can be avoided using CSA. Additionally, our approach can be extended to cases involving INT4 and INT2 weights, where the speedup over the baseline would be higher. For example, one 32-bit register can contain eight INT4 weights, and if seven of them are zeros, then the USSA will take a single clock cycle, whereas the baseline will take eight clock cycles.

\begin{figure}[b]
    \centering
    \resizebox{0.45\textwidth}{!}{\input{figures/fig2_iccad}}
    \caption{Analytical and observed speedups for SSSA (Semi-Structured Sparsity Accelerator).}
    \label{fig::speedup_sssa}
\end{figure}
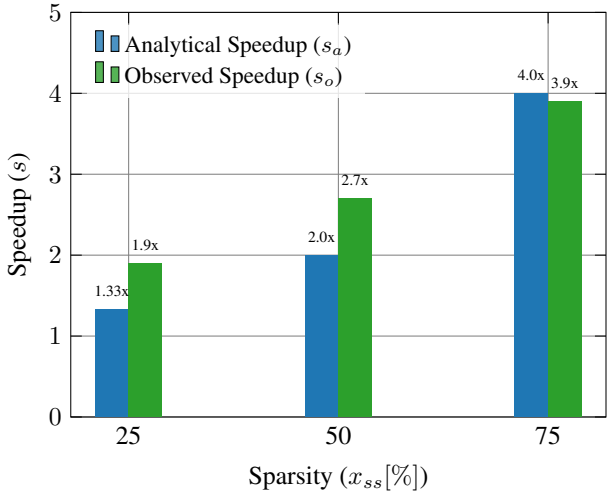

\subsection{Speedup for SSSA}\label{eval:sssa}

The SSSA (Semi-Structured Sparsity Accelerator) accelerates DNN models by exploiting semi-structured sparsity. The speedup is benchmarked against a baseline where every block of weights and inputs is processed, irrespective of their values. The analytical speedup ($s_a$) is calculated by the ratio of the total number of weights to the number of zero weights while the observed speedup ($s_o$) is measured by comparing the clock cycles required to process a convolutional layer in the baseline configuration versus the number of clock cycles needed by the SSSA.

\begin{table*}[ht]
\caption{Comparison of different methods for accelerating sparse DNNs.}
\small
\centering
\begin{tabular}{p{2.0cm}p{2.0cm}p{2.0cm}p{2.0cm}p{1.5cm}p{2cm}p{2cm}}
\hline
{\large\strut}\textbf{Method}          & \multicolumn{1}{c}{\textbf{Semi-Structured}} & \multicolumn{1}{c}{\textbf{Unstructured Sparsity}} & \multicolumn{1}{c}{\textbf{Sparsity Pattern}} & \multicolumn{1}{c}{\textbf{Speedup}} & \multicolumn{1}{c}{\textbf{Sparsity Ratios}} & \multicolumn{1}{c}{\textbf{Architecture}} \\ \hline
{\large\strut}Ours (USSA)              & \multicolumn{1}{c}{\xmark}                   & \multicolumn{1}{c}{\cmark}                         & \multicolumn{1}{c}{NA}                        & \multicolumn{1}{c}{2--3$\times$}     & \multicolumn{1}{c}{High}                     & \multicolumn{1}{c}{CPU+HW} \\
{\large\strut}Ours (SSSA)              & \multicolumn{1}{c}{\cmark}                   & \multicolumn{1}{c}{\xmark}                         & \multicolumn{1}{c}{4:4}                       & \multicolumn{1}{c}{2--4$\times$}     & \multicolumn{1}{c}{Low}                      & \multicolumn{1}{c}{CPU+HW} \\
{\large\strut}Ours (CSA)               & \multicolumn{1}{c}{\cmark}                   & \multicolumn{1}{c}{\cmark}                         & \multicolumn{1}{c}{4:4, random}               & \multicolumn{1}{c}{4--5$\times$}     & \multicolumn{1}{c}{Moderate}                 & \multicolumn{1}{c}{CPU+HW} \\ \hline
{\large\strut}IndexMAC~\cite{IndexMAC} & \multicolumn{1}{c}{\cmark}                   & \multicolumn{1}{c}{\xmark}                         & \multicolumn{1}{c}{2:4}                       & \multicolumn{1}{c}{2--3$\times$}     & \multicolumn{1}{c}{Moderate}                 & \multicolumn{1}{c}{CPU+HW} \\ \hline
{\large\strut}\citeauthor{Lu_etal}~\cite{Lu_etal}  & \multicolumn{1}{c}{NA}                       & \multicolumn{1}{c}{\cmark}                         & \multicolumn{1}{c}{Low}                       & \multicolumn{1}{c}{2.4--12.9$\times$}& \multicolumn{1}{c}{NA}                       & \multicolumn{1}{c}{HW} \\ \hline
\end{tabular}
\label{tab:pruning_comparison}
\end{table*}

\Cref{fig::speedup_sssa} provides a comparison of the two speedups. Note that the observed speedup ($s_o$) sometimes exceeds the analytical speedup ($s_a$) due to reduced overhead, as the accelerator bypasses entire blocks of zero values in a loop, eliminating unnecessary iterations. Speedups were measured for a convolutional layer.
It is to be noted that even higher speedups should be achievable for models with higher sparsity than considered in the paper.

\subsection{Speedup of DNN models using CSA}\label{eval:csa}
\begin{figure}[H]
    \centering
    \resizebox{0.45\textwidth}{!}{\input{figures/fig3_iccad}}
    \caption{Speedups of considered DNN models with CSA (Combined Sparsity Accelerator) for three different configurations of unstructured sparsity ($x_{us}$) and semi-structured sparsity ($x_{ss}$). Even higher speedups should be achievable for models with higher sparsity than considered in the paper.}
    \label{fig::iccad3}
\end{figure}
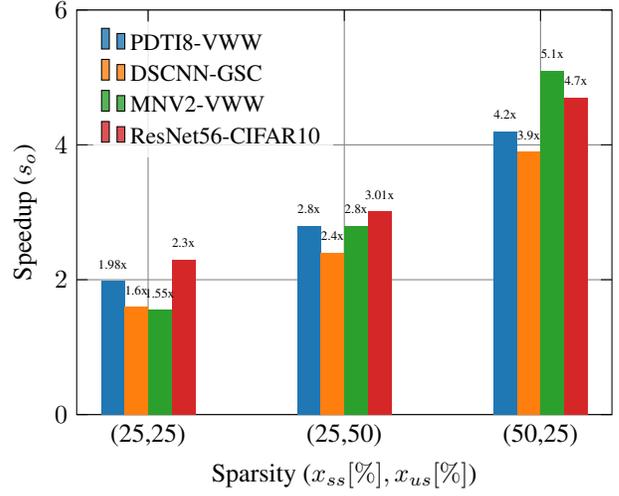

The Combined Sparsity Accelerator (CSA) integrates the functionalities of both prior designs. Observed speedups for four DNN models using the CSA are presented in \Cref{fig::iccad3}.

\subsection{Impact of Losing One Information Bit}\label{ssec:sacrifice}
In our design for semi-structured pruning, one bit per weight is used to encode sparsity information. Therefore, the model performance loss from sacrificing one bit needs to be analyzed.
In our experiments, we found that effectively reducing the precision from INT8 to INT7 for each of the three considered applications did not impact the accuracy. Concrete results are shown in \Cref{tab:1bit}.

\begin{table}[H]
\caption{Accuracy comparison between INT8 and INT7 precision.}
\centering
\small
\begin{tabular}{lcc}
\hline
{\large\strut}\textbf{Model-Dataset}     & \textbf{Accuracy (INT8)} & \textbf{Accuracy (INT7)} \\ \hline
{\large\strut}ResNet-56 on CIFAR10       &  93.51\%                 &    93.53\%                           \\
{\large\strut}MobileNetV2 on VWW         & 91.53\%                  &     91.42\%                          \\
{\large\strut}DSCNN on GSC               &  95.17\%                 &     95.10\%                          \\ \hline
\end{tabular}
\label{tab:1bit}
\end{table}

Our results are consistent with observations in previous works~\cite{int7_scale}, where it was found that quantizing to INT7 weights does not incur any noticeable loss in accuracy and may, in some cases, increase the performance on the test dataset. A slight increase in some cases may occur due to better generalization on the test dataset.

\vspace{-0.1cm}

\subsection{Comparison with the State-of-the-Art}

Though not easily comparable, we selected two other works for comparison: IndexMAC~\cite{IndexMAC} and a fully parallel sparse convolution accelerator by \citeauthor{Lu_etal}~\cite{Lu_etal}. \Cref{tab:pruning_comparison} summarizes the comparison.
The approach by \citeauthor{Lu_etal} employs a fully hardware-parallel architecture, while IndexMAC is a CPU architecture with instruction extensions added to a RISC-V core to accelerate semi-structured pruning. The tradeoff between a full hardware accelerator and a CPU extension is that the former offers higher throughput and lower latency due to the simultaneous processing of operations. In contrast, the latter is more resource-efficient, utilizing hardware resources more effectively by reusing them across multiple operations.

IndexMAC employs specific sparsity patterns, such as 1:4 or 2:4 pruning. This sparsity pattern lies \enquote{mid-way} between the two of our designs. Note that our unstructured pruning approach imposes no constraints on sparsity patterns. This leads to higher levels of sparsity. Our semi-structured pruning approach demands that a block of four weights be zero, which we refer to as a 4:4 sparsity pattern.

In terms of speedup, our unstructured pruning design (USSA) is superior to the IndexMAC approach for the reason that it provides similar speedup for similar sparsity but imposes no constraint on the specific sparsity pattern, leading to higher pruneability.

Moreover, our semi-structured pruning approach (SSSA) is both faster and simpler due to our novel lookahead encoding. By reserving one bit per weight for lookahead information, we efficiently skip blocks with zero weights, which accelerates the processing of DNNs. This novel method of leveraging embedded information within DNN weights to enhance acceleration is, to the best of our knowledge, not found in previous work. Another key aspect is the combination of semi-structured and unstructured pruning in a single design. This hybrid approach offers the best of both of our designs.

\vspace{-0.5mm}
\subsection{Resource Usage}\label{sec:ResourceUsage}

We provide FPGA resource usage for our designs for accelerating semi-structured and unstructured pruning in \Cref{tab:resourceUsageCombined}.
The resource usage was obtained for a Xilinx XC7A35T FPGA comprising 33,280 logic cells and 90 DSP slices. In both cases, the increase in LUT utilization is less than 4\%, and the usage of flip-flops (FF) is around 6\% while one additional DSPs are used. The two designs may be used independently or merged to accelerate both unstructured and semi-structured sparsity simultaneously at the cost of slightly more FPGA resource utilization. The combined design utilizes an additional 4.39\% more LUTs, 8.23\% more FFs, and two additional DSPs.
The standard clock frequency of 100~MHz of the RISC-V-based system (LiteX SoC~\cite{CFU-Playground}) was considered throughout all our experiments.
The addition of our proposed CFUs did not affect the frequency.
Note that if a CFU design would be on the critical path, there is the option to pipeline the CFU datapath, as CFU Playground allows for multicycle custom instructions.

\begin{table}[H]
\caption{FPGA resource usage comparison for different sparsity accelerator CFUs.}
\label{tab:resourceUsageCombined}
\centering
\small 
\begin{tabular}{lcccc}
\hline
{\large\strut}\textbf{Design Type} & \multicolumn{2}{c}{\textbf{RISC-V}} & \textbf{Cost Increment [\%]} \\
                      &  w/o CFU &  with CFU &  \\
\hline
{\large\strut}\textbf{USSA} \\
LUTs                  & 2,482 & 2,516 & 1.36\%  \\
Slice FF              & 1,470 & 1,563 & 6.32\%  \\
BRAMs                 & 9     & 9     & 0\%     \\
DSPs                  & 4     & 5     & 25\%    \\
\hline
{\large\strut}\textbf{SSSA} \\
LUTs                  & 2,473 & 2,568 & 3.84\%  \\
Slice FF              & 1,481 & 1,578 & 6.55\% \\
BRAMs                 & 9     & 9     & 0\%     \\
DSPs                  & 4     & 5     & 25\%    \\
\hline
{\large\strut}\textbf{CSA} \\
LUTs                  & 2,459 & 2,567 & 4.39\%  \\
Slice FF              & 1,470 & 1,591 & 8.23\% \\
BRAMs                 & 9     & 9     & 0\%     \\
DSPs                  & 4     & 6     & 50\%    \\
\hline
\end{tabular}
\end{table}
\vspace{-1mm}

\section{Conclusion and Future Work}
In this paper, we proposed novel RISC-V extensions aimed at accelerating DNNs by exploiting both unstructured and semi-structured sparsity. Our designs are customizable to fit various use cases and can benefit from FPGA reconfigurability. The novel aspect of our design to accelerate semi-structured pruning (SSSA) is the lookahead approach that encodes sparsity information into DNN weights by using the last bit of each INT8 weight. This strategy simplifies the design and requires minimal extra FPGA resources.
The novel aspect of our design to accelerate unstructured sparsity (USSA) is a variable-cycle sequential multiply-accumulate unit. We then combined both designs and achieved up to 5$\times$ speedup for selected DNN-based applications, with the potential for more significant speedups in other scenarios.

In order to overcome the limitation imposed by the register-to-register (64 bit) CFU-CPU interface, we also plan to investigate co-processor architectures in the future.

%% file: figures/fig1_iccad.tex
\begin{tikzpicture}

\definecolor{darkgray176}{RGB}{176,176,176}
\definecolor{lightgray204}{RGB}{204,204,204}

\begin{axis}[
    legend cell align={left},
    legend style={fill opacity=0.8, draw opacity=1, text opacity=1, draw=lightgray204, at={(0.01,0.99)}, anchor=north west, font=\small},
    width=\columnwidth,
    height=.7\columnwidth,
    ylabel near ticks,
    tick pos=left,
    xlabel={Sparsity ($x_{us}$[\%])},
    ylabel={Speedup (s)},
    xmin=-0, xmax=1,
    ymin=1, ymax=5,
    xmajorgrids, ymajorgrids, 
    grid style={darkgray176}, 
    xtick style={color=black},
    ytick style={color=black},
    yticklabel style={color=black}, 
    tick label style={}, 
    ytick={0,1,2,3,4}, 
    yticklabels={0, 1, 2, 3, 4}, 
    xticklabel={\pgfmathparse{\tick*100}\pgfmathprintnumber{\pgfmathresult}} 
]

\addplot [semithick, black]
table {%
0 1
0.02 1.02040816326531
0.04 1.04166666666667
0.06 1.06382978723404
0.08 1.08695652173913
0.1 1.11111111111111
0.12 1.13636363636364
0.14 1.16279069767442
0.16 1.19047619047619
0.18 1.21951219512195
0.2 1.25
0.22 1.28205128205128
0.24 1.31578947368421
0.26 1.35135135135135
0.28 1.38888888888889
0.3 1.42857142857143
0.32 1.47058823529412
0.34 1.51515151515152
0.36 1.5625
0.38 1.61290322580645
0.4 1.66666666666667
0.42 1.72413793103448
0.44 1.78571428571429
0.46 1.85185185185185
0.48 1.92307692307692
0.5 2
0.52 2.08333333333333
0.54 2.17391304347826
0.56 2.27272727272727
0.58 2.38095238095238
0.6 2.5
0.62 2.63157894736842
0.64 2.77777777777778
0.66 2.94117647058824
0.68 3.125
0.7 3.33333333333333
0.72 3.57142857142857
0.74 3.84615384615385
0.76 4.16666666666667
0.78 4.54545454545455
0.8 5
0.82 5.55555555555556
0.84 6.25
0.86 7.14285714285714
0.88 8.33333333333333
0.9 10
0.92 12.5
0.94 16.6666666666667
0.96 25
0.98 50
1 inf
};
\addlegendentry{Analytical Speedup ($s_a$)}

\addplot [semithick, red, dashed] 
table {%
0 1
0.02 1.020408121616
0.04 1.04166597222269
0.06 1.06382612042912
0.08 1.08694442357511
0.1 1.1110802477709
0.12 1.13629669815814
0.14 1.16266085819904
0.16 1.19024403621084
0.18 1.21912201660727
0.2 1.24937531234383
0.22 1.28108941384061
0.24 1.31435502811248
0.26 1.34926829989658
0.28 1.38593100414138
0.3 1.42445069620028
0.32 1.46494080232815
0.34 1.50752062846002
0.36 1.55231525958188
0.38 1.59945531508918
0.4 1.6490765171504
0.42 1.7013190190249
0.44 1.75632642830085
0.46 1.81424444592648
0.48 1.87521902558219
0.5 1.93939393939394
0.52 2.0069076154598
0.54 2.07788909075812
0.56 2.1524529008866
0.58 2.23069270771385
0.6 2.31267345050879
0.62 2.3984218000808
0.64 2.48791470552644
0.66 2.5810658583732
0.68 2.67770997102075
0.7 2.7775848899382
0.72 2.88031175572744
0.74 2.98537369892936
0.76 3.09209393632958
0.78 3.19961461281911
0.8 3.30687830687831
0.82 3.41261474405185
0.84 3.51533586362522
0.86 3.61334283208897
0.88 3.7047487172678
0.9 3.78752012120064
0.92 3.8595399181407
0.94 3.91869123236726
0.96 3.96295995698445
0.98 3.99054973972836
1 4
};
\addlegendentry{Observed Speedup ($s_o$)}

\end{axis}

\end{tikzpicture}

%% file: figures/fig2_iccad.tex
\begin{tikzpicture}

\definecolor{forestgreen4416044}{RGB}{44,160,44}
\definecolor{gray}{RGB}{128,128,128}
\definecolor{steelblue31119180}{RGB}{31,119,180}

\begin{axis}[
width=\columnwidth,height=.8\columnwidth,
    ylabel near ticks,
legend cell align={left},
legend style={
  fill opacity=0.8,
  draw opacity=1,
  text opacity=1,
  at={(0.03,0.97)},
  anchor=north west,
  draw=none,
  font=\small
},
tick pos=both,
x grid style={gray},
xlabel={Sparsity ($x_{ss}[\%]$)},
xmajorgrids,
xmin=18.1, xmax=81.9,
xtick style={color=black},
xtick={25,50,75},
xticklabels={25,50,75},
y grid style={gray},
ylabel={Speedup ($s$)},
ymajorgrids,
ymin=0, ymax=5,
ytick style={color=black}
]
\draw[draw=none,fill=steelblue31119180] (axis cs:21,0) rectangle (axis cs:25,1.33333333333333);
\addlegendimage{ybar,ybar legend,draw=none,fill=steelblue31119180}
\addlegendentry{Analytical Speedup ($s_a$)}

\draw[draw=none,fill=steelblue31119180] (axis cs:46,0) rectangle (axis cs:50,2);
\draw[draw=none,fill=steelblue31119180] (axis cs:71,0) rectangle (axis cs:75,4);
\draw[draw=none,fill=forestgreen4416044] (axis cs:25,0) rectangle (axis cs:29,1.9);
\addlegendimage{ybar,ybar legend,draw=none,fill=forestgreen4416044}
\addlegendentry{Observed Speedup ($s_o$)}

\draw[draw=none,fill=forestgreen4416044] (axis cs:50,0) rectangle (axis cs:54,2.7);
\draw[draw=none,fill=forestgreen4416044] (axis cs:75,0) rectangle (axis cs:79,3.9);
\draw (axis cs:23,1.33333333333333) ++(0pt,3pt) node[
  scale=0.6,
  anchor=south,
  text=black,
  rotate=0.0
]{1.33x};
\draw (axis cs:48,2) ++(0pt,3pt) node[
  scale=0.6,
  anchor=south,
  text=black,
  rotate=0.0
]{2.0x};
\draw (axis cs:73,4) ++(0pt,3pt) node[
  scale=0.6,
  anchor=south,
  text=black,
  rotate=0.0
]{4.0x};
\draw (axis cs:27,1.9) ++(0pt,3pt) node[
  scale=0.6,
  anchor=south,
  text=black,
  rotate=0.0
]{1.9x};
\draw (axis cs:52,2.7) ++(0pt,3pt) node[
  scale=0.6,
  anchor=south,
  text=black,
  rotate=0.0
]{2.7x};
\draw (axis cs:77,3.9) ++(0pt,3pt) node[
  scale=0.6,
  anchor=south,
  text=black,
  rotate=0.0
]{3.9x};
\end{axis}

\end{tikzpicture}

%% file: figures/fig3_iccad.tex
\begin{tikzpicture}

\definecolor{crimson2143940}{RGB}{214,39,40}
\definecolor{darkorange25512714}{RGB}{255,127,14}
\definecolor{forestgreen4416044}{RGB}{44,160,44}
\definecolor{gray}{RGB}{128,128,128}
\definecolor{steelblue31119180}{RGB}{31,119,180}

\begin{axis}[
width=\columnwidth,height=.8\columnwidth,
    ylabel near ticks,
legend cell align={left},
legend style={
  fill opacity=0.8,
  draw opacity=1,
  text opacity=1,
  at={(0.03,0.97)},
  anchor=north west,
  draw=none,
  font=\small
},
tick pos=both,
x grid style={gray},
xlabel={Sparsity ($x_{ss}[\%], x_{us}[\%]$)},
xmajorgrids,
xmin=15.9, xmax=84.1,
xtick style={color=black},
xtick={25,50,75},
  xticklabels={(25,25) ,(25,50), (50,25)},
y grid style={gray},
ylabel={Speedup ($s_o$)},
ymajorgrids,
ymin=0, ymax=6,
ytick style={color=black}
]
\draw[draw=none,fill=steelblue31119180] (axis cs:19,0) rectangle (axis cs:22,1.98);
\addlegendimage{ybar,ybar legend,draw=none,fill=steelblue31119180}
\addlegendentry{PDTI8-VWW}

\draw[draw=none,fill=steelblue31119180] (axis cs:44,0) rectangle (axis cs:47,2.8);
\draw[draw=none,fill=steelblue31119180] (axis cs:69,0) rectangle (axis cs:72,4.2);
\draw[draw=none,fill=darkorange25512714] (axis cs:22,0) rectangle (axis cs:25,1.6);
\addlegendimage{ybar,ybar legend,draw=none,fill=darkorange25512714}
\addlegendentry{DSCNN-GSC}

\draw[draw=none,fill=darkorange25512714] (axis cs:47,0) rectangle (axis cs:50,2.4);
\draw[draw=none,fill=darkorange25512714] (axis cs:72,0) rectangle (axis cs:75,3.9);
\draw[draw=none,fill=forestgreen4416044] (axis cs:25,0) rectangle (axis cs:28,1.55);
\addlegendimage{ybar,ybar legend,draw=none,fill=forestgreen4416044}
\addlegendentry{MNV2-VWW}

\draw[draw=none,fill=forestgreen4416044] (axis cs:50,0) rectangle (axis cs:53,2.8);
\draw[draw=none,fill=forestgreen4416044] (axis cs:75,0) rectangle (axis cs:78,5.1);
\draw[draw=none,fill=crimson2143940] (axis cs:28,0) rectangle (axis cs:31,2.3);
\addlegendimage{ybar,ybar legend,draw=none,fill=crimson2143940}
\addlegendentry{ResNet56-CIFAR10}

\draw[draw=none,fill=crimson2143940] (axis cs:53,0) rectangle (axis cs:56,3.01);
\draw[draw=none,fill=crimson2143940] (axis cs:78,0) rectangle (axis cs:81,4.7);
\draw (axis cs:20.5,1.98) ++(0pt,3pt) node[
  scale=0.5,
  anchor=south,
  text=black,
  rotate=0.0
]{1.98x};
\draw (axis cs:45.5,2.8) ++(0pt,3pt) node[
  scale=0.5,
  anchor=south,
  text=black,
  rotate=0.0
]{2.8x};
\draw (axis cs:70.5,4.2) ++(0pt,3pt) node[
  scale=0.5,
  anchor=south,
  text=black,
  rotate=0.0
]{4.2x};
\draw (axis cs:23.5,1.6) ++(0pt,3pt) node[
  scale=0.5,
  anchor=south,
  text=black,
  rotate=0.0
]{1.6x};
\draw (axis cs:48.5,2.4) ++(0pt,3pt) node[
  scale=0.5,
  anchor=south,
  text=black,
  rotate=0.0
]{2.4x};
\draw (axis cs:73.5,3.9) ++(0pt,3pt) node[
  scale=0.5,
  anchor=south,
  text=black,
  rotate=0.0
]{3.9x};
\draw (axis cs:26.5,1.55) ++(0pt,3pt) node[
  scale=0.5,
  anchor=south,
  text=black,
  rotate=0.0
]{1.55x};
\draw (axis cs:51.5,2.8) ++(0pt,3pt) node[
  scale=0.5,
  anchor=south,
  text=black,
  rotate=0.0
]{2.8x};
\draw (axis cs:76.5,5.1) ++(0pt,3pt) node[
  scale=0.5,
  anchor=south,
  text=black,
  rotate=0.0
]{5.1x};
\draw (axis cs:29.5,2.3) ++(0pt,3pt) node[
  scale=0.5,
  anchor=south,
  text=black,
  rotate=0.0
]{2.3x};
\draw (axis cs:54.5,3.01) ++(0pt,3pt) node[
  scale=0.5,
  anchor=south,
  text=black,
  rotate=0.0
]{3.01x};
\draw (axis cs:79.5,4.7) ++(0pt,3pt) node[
  scale=0.5,
  anchor=south,
  text=black,
  rotate=0.0
]{4.7x};
\end{axis}

\end{tikzpicture}